\documentclass[sigconf]{acmart}
\AtBeginDocument{%
  \providecommand\BibTeX{{%
    \normalfont B\kern-0.5em{\scshape i\kern-0.25em b}\kern-0.8em\TeX}}}

\copyrightyear{2020}
\acmYear{2020}
\setcopyright{acmlicensed}\acmConference[FIRE '20]{Forum for Information
Retrieval Evaluation}{December 16--20, 2020}{Hyderabad, India}
\acmBooktitle{Forum for Information Retrieval Evaluation (FIRE '20), December
16--20, 2020, Hyderabad, India}
\acmPrice{15.00}
\acmDOI{10.1145/3441501.3441541}
\acmISBN{978-1-4503-8978-5/20/12}

\addtocounter{page}{36}
\begin{document}

\title{UrduFake@FIRE2020: Shared Track on \\ Fake News Identification in Urdu}

\author{Maaz Amjad}
\email{maazamjad@phystech.edu}
\affiliation{%
  \institution{Center for Computing Research (CIC), Instituto Polit\'ecnico Nacional}
  \city{Mexico City}
  \country{Mexico}
}

\author{Grigori Sidorov}
\affiliation{%
  \institution{Center for Computing Research (CIC), Instituto Polit\'ecnico Nacional}
  \city{Mexico City}
  \country{Mexico}}
\email{sidorov@cic.ipn.mx}

\author{Alisa Zhila}
\affiliation{%
  \institution{Independent Researcher}
  \city{}
  \country{United States}
}

\author{Alexander Gelbukh}
\affiliation{%
  \institution{Center for Computing Research (CIC), Instituto Polit\'ecnico Nacional}
  \city{Mexico City}
  \country{Mexico}}
\email{gelbukh@gelbukh.com}

\author{Paolo Rosso}
\affiliation{%
  \institution{PRHLT Research Center, \\Universitat Polit\'ecnica de Val\'encia}
  \country{Spain}}
\email{prosso@dsic.upv.es}

\renewcommand{\shortauthors}{Amjad, Sidorov et al.}

\begin{abstract}
This paper gives the overview of the first shared task at FIRE 2020 on fake news detection in the Urdu language. This is a binary classification task in which the goal is to identify fake news using a dataset composed of 900 annotated news articles for training and 400 news articles for testing. The dataset contains news in five domains: (i) Health, (ii) Sports, (iii) Showbiz, (iv) Technology, and (v) Business. 42 teams from 6 different countries (India, China, Egypt, Germany, Pakistan, and the UK) registered for the task. 9 teams submitted their experimental results. The participants used various machine learning methods ranging from feature-based traditional machine learning to neural network techniques. The best performing system achieved an F-score value of 0.90, showing that the BERT-based approach outperforms other machine learning classifiers.
\end{abstract}




\maketitle

\section{Introduction}

Automatic detection of fake news is an essential task, because fake news are often presented to the global audience to sway its opinion. This variation of public opinion can be manipulated for various objectives such as a political campaign or induction to a war, etc. Although fake news causes excitement or fear and spurs action, however, the propagation of fake news brings devastating and havoc impact in our society. For example, BBC\footnote{https://www.bbc.com/news/world-asia-india-53165436} reported in a recent study that fake news had severe consequences for minority communities and some businesses in India. The study highlighted that some Muslims were victimized as responsible for spreading the coronavirus, and such fake news led the calls for an economic boycott of Muslim business. Likewise, in a different study, BBC\footnote{https://www.bbc.com/news/world-asia-54649302} reported that some Indian sites claimed a civil war had broken out in one of the biggest city of Pakistan, known as Karachi, which eventually turned out a fake news. Therefore, combating and controlling the speed of fake news propagation by automatic detection is a vital task to support vulnerable communities and maintain a trustworthy news ecosystem.

Automatic fake news detection task received the attention of many researchers, especially after the US 2016 presidential election, where fake news were widely used as a political instrument during the campaign.
For example, recent overview papers reported scientific issues associated with fake news detection \cite{1,2}. This shared task is greatly inspired by three previous evaluation forums,  PAN@CLEF 2020 \cite{3}, MediaEval\footnote{https://multimediaeval.github.io/editions/2020/tasks/fakenews/}, and RumourEval \cite{4}.  Significant shared tasks in several languages, in particular for English\footnote{http://www.fakenewschallenge.org/}$^,$\footnote{https://pan.webis.de/clef20/pan20-web/author-profiling.html}, Spanish\footnote{https://pan.webis.de/clef20/pan20-web/author-profiling.html} and Arabic \cite{5} were designed. Usually, the studies focus on the resource-rich languages, in terms of size of annotated datasets and availability of NLP tools, such as English \cite{7}, Spanish \cite{8}, German \cite{9} and Chinese \cite{10}. As the spread of fake news is still growing on the internet and digital media, therefore technological support to identify fake content is necessary for all languages, including low resource languages like Urdu.

This paper provides a brief description of the UrduFake track at FIRE 2020. The study \cite{6} provides a detail explanation of the problem definition and the participating team systems and describes the dataset, which we created for the shared task and made available to the research community. There is a growing need for research on the classification of fake news. Specifically, supervised machine learning algorithms based on a set of annotated training examples (news articles) can help to detect fake news. Keeping this in mind, we organize the 1st competition to address the current challenges with fake news detection for a low resource language (Urdu language, in our case). This competition also aims to encourage further development of resources in the form of annotated datasets and different digital solutions to combat the spread of fake news on the Web and digital media platforms.

\section{Task Description}
This is a binary classification task, when the goal is to assign a label (fake or real news) for given news articles written in the Urdu. In a different study \cite{11}, we presented the definition of fake news. The mathematical formulation of fake news article and fake news detection is as follows:

\begin{itemize}
\item {``\verb|Fake News Article|''}: A fake news is a factually incorrect news article, which provides factually incorrect information with the intention to deceive a reader making him to  believe it is true news.

\item {``\verb|Fake News Detection|'' }: For a given news article (unannotated), call as  $\alpha$, where $\alpha$  $\in$ $N$ ($\alpha$  is a news article out of $N$ news articles), an automatic fake news detection algorithm assigns a score $S(\alpha)$ $\in$ [0, 1] indicating the extent to which $S(\alpha)$ is considered as fake news article. For instance, if $S(\widehat{\alpha})$ > $S(\alpha)$, then it can be inferred that $\widehat{\alpha}$ is more likely to be a fake news article. A threshold $\gamma$ can be defined, such that the prediction function $F$ : $N$ $\rightarrow$ 
[not fake, fake] is:

\begin{align*}
F(N) &=  \begin{cases}
fake, \,\,\,\,\,\,if \,\,\,\,\ S(\alpha) \in \gamma), \\
not fake, \,\,\,\,\,\,\,\,\  \textrm{otherwise}.
\end{cases}
\end{align*}
\end{itemize}

\section{Data Collection and Annotation }
For this competition, we created a dataset for automatic fake news detection task in Urdu. This dataset contains news articles in five domains: (i) Business, (ii) Health, (iii) Showbiz (entertainment), (iv) Sports, and (v) Technology. This section briefly describes the collection procedure of real and fake news. The dataset is publicly available to use for academic research\footnote{https://github.com/UrduFake/urdufake2020eval.git}.

At the collection phase, thousands of news articles were crawled  from multiple national and reliable international mainstream media agencies. To collect news articles from numerous online sources, a Python library Newspaper\footnote{https://newspaper.readthedocs.io/en/latest/} was used as a web scraper to extract and curate the content of news articles from multiple national and international newspaper web pages. This library automatically discards irrelevant information such as HTML tags, author’s name, location of the publisher, noisy texts and images, and advertisements.

\begin{itemize}
\item \textit{\textbf{Real News Collection for Training and Testing Sets}}:\\ 
To manually collect and annotate news articles as a real news article, numerous articles were retrieved from various national and international mainstream news channels from January 2018 to December 2018. The list of the news agencies used to crawl real news and the procedure of verification for news authenticity is described in \cite{11}. A news article qualifies to be a real news only if it is published in a reliable newspaper and information such as place of the event, image, and date can be verified through prominent news agencies. We read the complete news article to check whether a news article correlates with the title and its content before annotation. Similarly, the same procedure was followed to collect and annotate the real news articles for the testing dataset. Nonetheless, all the news for the testing dataset were retrieved from January 2019 to June 2020.\\

\item \textit{\textbf{Professional  Crowdsourcing  to  Collect  Fake  News  for  Testing and Training Sets}}:\\ 
The collection and annotation of fake news articles, which correspond to the real news articles is a difficult task. Therefore,  we used professional journalist services from various news agencies in Pakistan (Dawn news, Express news, etc.). They were asked to write fake news stories that correspond to the original real news articles. The journalists wrote fake news for both training and testing datasets. This approach is similar to fake news dataset development for English language \cite{7}, where the authors as well adopted professional crowdsourcing to collect fake news.
\end{itemize}

Table {\ref{tab:freqq}} describes the distribution of the news articles in the dataset. 

\begin{table}[!htbp]
  \caption{Distribution of the news articles.}
  \label{tab:freqq}
  \begin{tabular}{cccl}
    \toprule
    &Real News  & Fake News  & Total\\
    \midrule
    \textbf{Train} &  500  & 400  & 900\\
    \textbf{Test}  &  250  & 150  & 400\\
  \bottomrule
 \textbf{ Total}   & 750   & 550  & 1300 \\
  \bottomrule
\end{tabular}
\end{table}

\section{Evaluation Metrics}
In this shared task, the task is to classify a news article as either fake or real news article. Initially, to develop and train automatic fake news detection systems, we released the training dataset for the participants. In the next stage, the test dataset was released to test and evaluate the performance of the system. Each participating team could submit only 3 different runs for evaluation.

The submitted systems were evaluated by comparing the labels predicted by the participants’ classifiers and the ground truth labels. For quantifying the classification performance, we employed the commonly used evaluation metrics: Precision (P), Recall (R), Accuracy, and two F1-scores (F1-score for each class and F1-macro). The F1-score has many variants like weighted F1, F1-macro or F1-micro. We calculated F1\textsubscript{real} to predict the label of the “real” class, F1\textsubscript{fake} to predict the label of the “fake” class out of all news, and F1-macro.

Fake news detection classification task suffers from class imbalance. The distribution of class labels is often unbalanced in datasets, which also happens in our case (i.e., we have more real news than fake news articles). Therefore, to accommodate the skew towards the real class, which dominates (it has more samples than the fake news class), we used the macro-averaged F1-macro, which is the average of F1\textsubscript{real} and F1\textsubscript{fake}. F1-macro does not use weights for the aggregation. However, F1-macro  penalizes when a system does not perform well for the minority classes. Although weighted-F1 is also calculated independently for each class, but when the weighted-F1 of both classes is summed up, it gives more weight to the majority class. Therefore, we only report F1-macro.

\section{Baseline Systems}

We provided three baseline systems with the goal that their performance could serve as reference points for qualitative evaluation of the submissions’ placement in the ranking. First, we provided the Random Baseline as the most basic and trivial baseline, which is expected to be ranked at the bottom with a more massive gap from the participating systems. Second, we provided the most traditional baseline: bag of words (BoW) model. It uses words as features and then apply a machine learning classifier. In this baseline, we used binary weighting scheme (i.e., a feature is present or not) with Logistic Regression classifier. For the third baseline, we provided the results of character bi-gram with tf-idf weighting scheme using Logistic Regression classifier, which achieved surprisingly good results. In addition, for the last two baselines, we tried five weighting schemes (tf-idf, logent, norm, binary, relative frequency) \cite{11} along with various classifiers such as Logistic Regression, SVM, Adaboost, Decision Tree, Random Forest, and Naive Bayes (we got the best results with Logistic Regression).

\section{Overview of the Submitted Approaches}

This section gives a brief overview of the systems submitted to this competition. 42 teams registered for participation, from which 9 teams submitted their runs. Registered participants were from 6 different countries (India, Pakistan, China, Egypt, Germany, and the UK). Participation of different teams from multiple countries confirms the importance of this task. The team members came from various types of organizations: universities, research centres, and industry. Table {\ref{tab:freq}} describes the submitted approaches.

\begin{table}[!htb]
  \caption{Approaches used by the participating systems.}
  \resizebox{\columnwidth}{!}
  {%
   \label{tab:freq}
  \begin{tabular}{ccccl}
    \toprule
System/Team Name & Feature Type & Feature Weighting Scheme & Classifying algorithm & NN-based\\
    \midrule
    
BERT 4EVER &   context embedding BERT &  BERT   &  CharCNN-Roberta   & Yes\\
Character bi-gram (baseline) &  char  bi-grams & TF-IDF   &  Logistic Regression  & No\\
BoW (baseline)&  word uni-grams & Binary   &  Logistic Regression  & No\\
CNLP-NITS &   $N/A$   &  embedding   &  XLNet pre-trained model     & Yes\\
NITP\_AI\_NLP &  char 1-3 grams   &  TF-IDF    &  Dense Neural Network   & Yes \\
Chanchal\_Suman &  $N/A$    & embeddings    & Bi-directional GRU model      & Yes \\
MUCS & mix of char and word n-grams    & embeddings    & ULMFiT model    & Yes\\
SSNCSE\_NLP & char n-gram & TFIDF, fastText, word2vec& RF, Adaboost, MLP, SVM    & Yes\\

  \bottomrule
\end{tabular}
}
\end{table}

\section{Results and Discussion}

Table {\ref{tab:overlapping2}} describes the results of the best runs of the submitted systems.   

\begin{table*}[ht]
\caption{Participants’ best run scores. }
\label{tab:overlapping2}
\begin{tabular}{ccccccccc}
\hline\noalign{\smallskip}
{\textbf{Team names}}& \multicolumn{3}{c}{\textbf{Fake Class}} & \multicolumn{3}{c}{\textbf{Real Class}} & \multicolumn{1}{c}{\textbf{F1-macro}} & \multicolumn{1}{c}{\textbf{Accuracy}}\\& \textbf{Precision} & \textbf{Recall}&  \textbf{F1\textsubscript{Fake}} &\textbf{Precision} & \textbf{Recall} & \textbf{F1\textsubscript{Real}} \\
	\noalign{\smallskip}\hline\noalign{\smallskip}
	\textbf{BERT 4EVER} & 0.890 & 0.860 & 0.874 & 0.918 & 0.936 & 0.926 & 0.900 & 0.908\\
	\noalign{\smallskip}\hline\noalign{\smallskip}
	\textbf{\textit{Character bi-gram (baseline)}} & 0.833 & 0.900 & 0.863 & 0.936 & 0.892 & 0.913 & 0.889& 0.895\\
	\noalign{\smallskip}\hline\noalign{\smallskip}
	\textbf{CNLP-NITS} & 0.836 & 0.713 & 0.769 & 0.842 & 0.916 & 0.877 & 0.823 & 0.840 \\
	\noalign{\smallskip}\hline\noalign{\smallskip}
	\textbf{NITP-AI-NLP} & 0.890 & 0.593 & 0.712 & 0.797 & 0.956 & 0.869 & 0.791 & 0.820 \\
	\noalign{\smallskip}\hline\noalign{\smallskip}
	\textbf{Chanchal-Suman} & 0.881 & 0.593 & 0.709 & 0.796 & 0.952 & 0.867 & 0.788 & 0.818 \\
	\noalign{\smallskip}\hline\noalign{\smallskip}
	\textbf{\textit{BoW (baseline)}} & 0.722 & 0.746 & 0.734 & 0.845 & 0.828 & 0.836 & 0.785 & 0.798 \\
	\noalign{\smallskip}\hline\noalign{\smallskip}
	\textbf{SSNNLP} & 0.709 & 0.733 & 0.721 & 0.837 & 0.820 & 0.828 & 0.774 & 0.787 \\			
    \noalign{\smallskip}\hline\noalign{\smallskip}
	\textbf{MUCS} & 0.783 & 0.627 & 0.696 & 0.800 & 0.896 & 0.845 & 0.770 & 0.795 \\
	\noalign{\smallskip}\hline\noalign{\smallskip}
	\textbf{CoDTeEM, NUST} & 0.771 & 0.607 & 0.679 & 0.791 & 0.892 & 0.838 & 0.758 & 0.785\\
	\noalign{\smallskip}\hline\noalign{\smallskip}
	\textbf{Rana Abdul Rehman} & 0.422 & 0.433 & 0.427 & 0.654 & 0.644 & 0.649 & 0.538 & 0.565 \\
	\noalign{\smallskip}\hline\noalign{\smallskip}
	\textit{\textbf{Random (baseline)}} &0.373 & 0.420 &0.395&0.623 & 0.576 & 0.599 & 0.497 &0.517\\
	\noalign{\smallskip}\hline\noalign{\smallskip}
	\textbf{Cyber Pilots} & 0.377 & 0.533 & 0.441 & 0.628 & 0.472 & 0.538 & 0.490 & 0.495\\
	\noalign{\smallskip}\hline\noalign{\smallskip}
\end{tabular}
\end{table*}

The participating team ($BERT 4EVER$) achieved the best F1-macro, Accuracy, as well as R\textsubscript{fake} (recall), F1\textsubscript{fake}, and P\textsubscript{real} (precision) scores. However, the baseline approach with character bi-grams and Logistic Regression achieved the second position in the shared task with just 1.1\% difference in F1-macro from BERT, which is quite an unexpected result. Explanation of this fact is a question for further research.

At this moment, it is hard to judge whether any of these approaches is ready to be applied “in the wild”. While the results of F1\textsubscript{real} and F1\textsubscript{fake} over 0.9 shown by the winning BERT 4EVER system are impressively high, the modest size of the provided training and testing dataset cannot guarantee the same performance on an arbitrary text input. To ensure the scalability of the presented approaches, more multifaceted research at a larger scale is needed. We see that one of the paths is a community-driven effort towards the increase of available resources and datasets in the Urdu language.

\section{Conclusion} 
This competition was aimed at supporting and encouraging researchers working in different NLP domains to develop robust technology to tackle and minimize the propagation of fake content in Urdu on the web. Note that Urdu is low resource language, i.e., no significant datasets or NLP tools are available. We presented the dataset for fake news detection for Urdu news articles, which contains 1,300 news. Real news were obtained from reliable sources and verified manually, while fake news were written by professional journalists following specific instructions. We described the approaches presented for this shared task. Some approaches were based on non-neural networks while other approaches were based on neural networks.  The best results were obtained by the approach based on BERT.  In future, we plan to make our corpus larger and more robust. In addition, it will be interesting to try transfer learning approaches, e.g., train a system to detect fake news on Spanish or English  and then test it on Urdu.

\begin{acks}
This competition was organized with partial support of National Council for Science and Technology (CONACYT) A1-S-47854, SIP-IPN 20200797 and 20200859 and CICLing conference. The work of the last author was partially funded by MICINN under the research project MISMIS-FAKEnHATE on MISinformation and MIScommunication in social media: FAKE news and HATE speech (PGC2018- 096212-B-C31).
\end{acks}

\bibliographystyle{ACM-Reference-Format}
\bibliography{main}


\begin{thebibliography}{11}


\ifx \showCODEN    \undefined \def \showCODEN     #1{\unskip}     \fi
\ifx \showDOI      \undefined \def \showDOI       #1{#1}\fi
\ifx \showISBNx    \undefined \def \showISBNx     #1{\unskip}     \fi
\ifx \showISBNxiii \undefined \def \showISBNxiii  #1{\unskip}     \fi
\ifx \showISSN     \undefined \def \showISSN      #1{\unskip}     \fi
\ifx \showLCCN     \undefined \def \showLCCN      #1{\unskip}     \fi
\ifx \shownote     \undefined \def \shownote      #1{#1}          \fi
\ifx \showarticletitle \undefined \def \showarticletitle #1{#1}   \fi
\ifx \showURL      \undefined \def \showURL       {\relax}        \fi
\providecommand\bibfield[2]{#2}
\providecommand\bibinfo[2]{#2}
\providecommand\natexlab[1]{#1}
\providecommand\showeprint[2][]{arXiv:#2}

\bibitem[\protect\citeauthoryear{Amjad, Sidorov, Zhila, Gelbukh, and
  Rosso}{Amjad et~al\mbox{.}}{2020a}]%
        {6}
\bibfield{author}{\bibinfo{person}{Maaz Amjad}, \bibinfo{person}{Grigori
  Sidorov}, \bibinfo{person}{Alisa Zhila}, \bibinfo{person}{Alexander Gelbukh},
  {and} \bibinfo{person}{Paolo Rosso}.} \bibinfo{year}{2020}\natexlab{a}.
\newblock \showarticletitle{Overview of the shared task on fake news detection
  in {Urdu} at {FIRE} 2020}.
\newblock \bibinfo{journal}{\emph{CEUR Workshop Proceedings}}
  (\bibinfo{year}{2020}).
\newblock
\newblock
\shownote{Working Notes of the Forum for Information Retrieval Evaluation (FIRE
  2020), Hyderabad, India.}


\bibitem[\protect\citeauthoryear{Amjad, Sidorov, Zhila, G\'{o}mez-Adorno,
  Voronkov, and Gelbukh}{Amjad et~al\mbox{.}}{2020b}]%
        {11}
\bibfield{author}{\bibinfo{person}{Maaz Amjad}, \bibinfo{person}{Grigori
  Sidorov}, \bibinfo{person}{Alisa Zhila}, \bibinfo{person}{Helena
  G\'{o}mez-Adorno}, \bibinfo{person}{Ilia Voronkov}, {and}
  \bibinfo{person}{Alexander Gelbukh}.} \bibinfo{year}{2020}\natexlab{b}.
\newblock \showarticletitle{Bend the {T}ruth: A benchmark dataset for fake news
  detection in {U}rdu and its evaluation}.
\newblock \bibinfo{journal}{\emph{Journal of Intelligent \& Fuzzy Systems}}
  \bibinfo{volume}{39}, \bibinfo{number}{2} (\bibinfo{year}{2020}),
  \bibinfo{pages}{2457--2469}.
\newblock
\urldef\tempurl%
\url{https://doi.org/10.3233/JIFS-179905}
\showDOI{\tempurl}


\bibitem[\protect\citeauthoryear{Bondielli and Marcelloni}{Bondielli and
  Marcelloni}{2019}]%
        {1}
\bibfield{author}{\bibinfo{person}{Alessandro Bondielli} {and}
  \bibinfo{person}{Francesco Marcelloni}.} \bibinfo{year}{2019}\natexlab{}.
\newblock \showarticletitle{A survey on fake news and rumour detection
  techniques}.
\newblock \bibinfo{journal}{\emph{Information Sciences}}  \bibinfo{volume}{497}
  (\bibinfo{year}{2019}), \bibinfo{pages}{38--55}.
\newblock


\bibitem[\protect\citeauthoryear{Gorrell, Kochkina, Liakata, Aker, Zubiaga,
  Bontcheva, and Derczynski}{Gorrell et~al\mbox{.}}{2019}]%
        {4}
\bibfield{author}{\bibinfo{person}{Genevieve Gorrell}, \bibinfo{person}{Elena
  Kochkina}, \bibinfo{person}{Maria Liakata}, \bibinfo{person}{Ahmet Aker},
  \bibinfo{person}{Arkaitz Zubiaga}, \bibinfo{person}{Kalina Bontcheva}, {and}
  \bibinfo{person}{Leon Derczynski}.} \bibinfo{year}{2019}\natexlab{}.
\newblock \showarticletitle{{S}em{E}val-2019 Task 7: {R}umour{E}val,
  determining rumour veracity and support for rumours}. In
  \bibinfo{booktitle}{\emph{Proceedings of the 13th International Workshop on
  Semantic Evaluation}}. \bibinfo{publisher}{Association for Computational
  Linguistics}, \bibinfo{address}{Minneapolis, Minnesota, USA},
  \bibinfo{pages}{845--854}.
\newblock
\urldef\tempurl%
\url{https://doi.org/10.18653/v1/S19-2147}
\showDOI{\tempurl}


\bibitem[\protect\citeauthoryear{Guo, Cao, Zhang, Shu, and Liu}{Guo
  et~al\mbox{.}}{2019}]%
        {10}
\bibfield{author}{\bibinfo{person}{Chuan Guo}, \bibinfo{person}{Juan Cao},
  \bibinfo{person}{Xueyao Zhang}, \bibinfo{person}{Kai Shu}, {and}
  \bibinfo{person}{Huan Liu}.} \bibinfo{year}{2019}\natexlab{}.
\newblock \showarticletitle{{DEAN}: Learning dual emotion for fake news
  detection on social media}.
\newblock \bibinfo{journal}{\emph{arXiv preprint arXiv:1903.01728}}
  (\bibinfo{year}{2019}).
\newblock


\bibitem[\protect\citeauthoryear{P{\'e}rez-Rosas, Kleinberg, Lefevre, and
  Mihalcea}{P{\'e}rez-Rosas et~al\mbox{.}}{2018}]%
        {7}
\bibfield{author}{\bibinfo{person}{Ver{\'o}nica P{\'e}rez-Rosas},
  \bibinfo{person}{Bennett Kleinberg}, \bibinfo{person}{Alexandra Lefevre},
  {and} \bibinfo{person}{Rada Mihalcea}.} \bibinfo{year}{2018}\natexlab{}.
\newblock \showarticletitle{Automatic detection of fake news}. In
  \bibinfo{booktitle}{\emph{Proceedings of the 27th International Conference on
  Computational Linguistics}}. \bibinfo{publisher}{Association for
  Computational Linguistics}, \bibinfo{address}{Santa Fe, New Mexico, USA},
  \bibinfo{pages}{3391--3401}.
\newblock
\urldef\tempurl%
\url{https://www.aclweb.org/anthology/C18-1287}
\showURL{%
\tempurl}


\bibitem[\protect\citeauthoryear{Posadas-Dur{\'a}n, G{\'o}mez-Adorno, Sidorov,
  and Escobar}{Posadas-Dur{\'a}n et~al\mbox{.}}{2019}]%
        {8}
\bibfield{author}{\bibinfo{person}{Juan-Pablo Posadas-Dur{\'a}n},
  \bibinfo{person}{Helena G{\'o}mez-Adorno}, \bibinfo{person}{Grigori Sidorov},
  {and} \bibinfo{person}{Jes{\'u}s Jaime~Moreno Escobar}.}
  \bibinfo{year}{2019}\natexlab{}.
\newblock \showarticletitle{Detection of fake news in a new corpus for the
  {S}panish language}.
\newblock \bibinfo{journal}{\emph{Journal of Intelligent \& Fuzzy Systems}}
  \bibinfo{volume}{36}, \bibinfo{number}{5} (\bibinfo{year}{2019}),
  \bibinfo{pages}{4869--4876}.
\newblock


\bibitem[\protect\citeauthoryear{Rangel, Giachanou, Ghanem, and Rosso}{Rangel
  et~al\mbox{.}}{2020}]%
        {3}
\bibfield{author}{\bibinfo{person}{Francisco Rangel},
  \bibinfo{person}{Anastasia Giachanou}, \bibinfo{person}{Bilal Ghanem}, {and}
  \bibinfo{person}{Paolo Rosso}.} \bibinfo{year}{2020}\natexlab{}.
\newblock \showarticletitle{Overview of the 8th author profiling task at {PAN}
  2020: profiling fake news spreaders on Twitter}.
\newblock \bibinfo{journal}{\emph{CEUR Workshop Proceedings}}
  \bibinfo{volume}{2696}.
\newblock


\bibitem[\protect\citeauthoryear{Rangel, Rosso, Charfi, Zaghouani, Ghanem, and
  S{\'a}nchez-Junquera}{Rangel et~al\mbox{.}}{2019}]%
        {5}
\bibfield{author}{\bibinfo{person}{Francisco Rangel}, \bibinfo{person}{Paolo
  Rosso}, \bibinfo{person}{Anis Charfi}, \bibinfo{person}{Wajdi Zaghouani},
  \bibinfo{person}{Bilal Ghanem}, {and} \bibinfo{person}{Javier
  S{\'a}nchez-Junquera}.} \bibinfo{year}{2019}\natexlab{}.
\newblock \showarticletitle{On the author profiling and deception detection in
  {A}rabic shared task at {FIRE}}, In \bibinfo{booktitle}{Proceedings of the
  11th Forum for Information Retrieval Evaluation}.
\newblock \bibinfo{journal}{\emph{CEUR Workshop Proceedings}}
  \bibinfo{volume}{2517}, \bibinfo{pages}{70--83}.
\newblock


\bibitem[\protect\citeauthoryear{Shu, Sliva, Wang, Tang, and Liu}{Shu
  et~al\mbox{.}}{2017}]%
        {2}
\bibfield{author}{\bibinfo{person}{Kai Shu}, \bibinfo{person}{Amy Sliva},
  \bibinfo{person}{Suhang Wang}, \bibinfo{person}{Jiliang Tang}, {and}
  \bibinfo{person}{Huan Liu}.} \bibinfo{year}{2017}\natexlab{}.
\newblock \showarticletitle{Fake news detection on social media: A data mining
  perspective}.
\newblock \bibinfo{journal}{\emph{ACM SIGKDD explorations newsletter}}
  \bibinfo{volume}{19}, \bibinfo{number}{1} (\bibinfo{year}{2017}),
  \bibinfo{pages}{22--36}.
\newblock


\bibitem[\protect\citeauthoryear{Vogel and Jiang}{Vogel and Jiang}{2019}]%
        {9}
\bibfield{author}{\bibinfo{person}{Inna Vogel} {and} \bibinfo{person}{Peter
  Jiang}.} \bibinfo{year}{2019}\natexlab{}.
\newblock \showarticletitle{Fake news detection with the new {G}erman dataset
  “{GermanFakeNC}”}. In \bibinfo{booktitle}{\emph{International Conference
  on Theory and Practice of Digital Libraries}}. Springer,
  \bibinfo{pages}{288--295}.
\newblock


\end{thebibliography}

\appendix

\end{document}